\documentclass[letterpaper, 10pt, conference]{ieeeconf}  % RA-L first submission (US Letter, conference)

\IEEEoverridecommandlockouts                              % needed for \thanks
\usepackage{graphicx}
\usepackage{amsmath,amssymb}
\usepackage{url}
\usepackage{xcolor}
\usepackage{algorithm}
\usepackage{algorithmic}

\title{\LARGE \bf
\textbf{H\textsuperscript{2}MARL}: Hierarchical Heterogeneous Multi-agent Reinforcement Learning for Humanoid Robot Control
}

\author{Xuemin Hu$^{1}$, Kaixin Li$^{1}$, Qingqing Xia$^{1}$, Yuxin Wang$^{1}$, Wei Xie$^{1}$, Bo Tang$^{2}$, and Long Chen$^{3}$% <-this % stops a space
\thanks{$^{1}$School of Artificial Intelligence, Hubei University, Wuhan, China. {\tt\small huxuemin2012@hubu.edu.cn}}%
\thanks{$^{2}$Worcester Polytechnic Institute, Worcester, MA, USA. {\tt\small btang1@wpi.edu}}%
\thanks{$^{3}$Institute of Automation, Chinese Academy of Sciences, Beijing, China. {\tt\small long.chen@ia.ac.cn}}%
}

\begin{document}

\maketitle
\thispagestyle{empty}
\pagestyle{empty}
\noindent{\small\textbf{\textit{Abstract}}---\,Coordinated whole-body control of humanoid robots remains challenging due to high-dimensional coupled dynamics and substantial physical heterogeneity across joint degrees of freedom (JDoFs). Existing multi-agent reinforcement learning methods either employ coarse agent partitions or directly apply generic heterogeneous-agent algorithms, while hierarchical controllers commonly optimize high-level planning and low-level execution independently, leading to uncoordinated motion. In this paper, we propose {H\textsuperscript{2}MARL}, a hierarchical heterogeneous multi-agent reinforcement learning framework with two key mechanisms. We first formulate humanoid whole-body control using a JDoF-level physical-heterogeneity-aware policy factorization, where each torque-actuated JDoF is conditioned on an explicit descriptor of its actuation, kinematic, and dynamic properties. Moreover, we develop a closed-loop policy–executor co-optimization mechanism that jointly adapts the high-level pose planner and the low-level joint executor using shared interaction trajectories, thereby reducing the distribution mismatch between planning and execution. %HAPPO and HST are used as the optimization and temporal-execution backbones, respectively. 
Extensive experiments in MuJoCo and Isaac Gym demonstrate that {H\textsuperscript{2}MARL} consistently improves locomotion stability, gait coordination, and motion quality over homogeneous, coarse-grained, and independently optimized alternatives.}

\par\vspace{1ex}

\section{Introduction}
Coordinated whole-body control (WBC) is a major challenge in achieving highly human-like movement for humanoid robot (HR) research~\cite{li2025humanoid}. The current mainstream methods are based on imitation learning and have achieved remarkable results ~\cite{tevet2025closd, fu2024humanplus}. However, these methods require a large amount of labeled data through motion capture systems or teleoperation, resulting in high training costs. Moreover, some marginal data, such as actions in complex and dangerous industrial and outer space operations, are difficult to collect, limiting the application of these methods~\cite{Hu2026Enhancing}. Reinforcement learning (RL)-based methods~\cite{hu2026Long} mainly involve single-agent RL methods, which can complete motions such as walking and running in simulators without a large amount of data and exhibit good coordination capabilities for HRs with low degrees of freedom (DoF)~\cite{schulman2017proximal,  radosavovic2024real}, but these methods struggle to achieve precise coordination across joints with numerous joints and high DoF. 

\begin{figure}[tb!]
    \centering
    \includegraphics[width=0.8\linewidth]{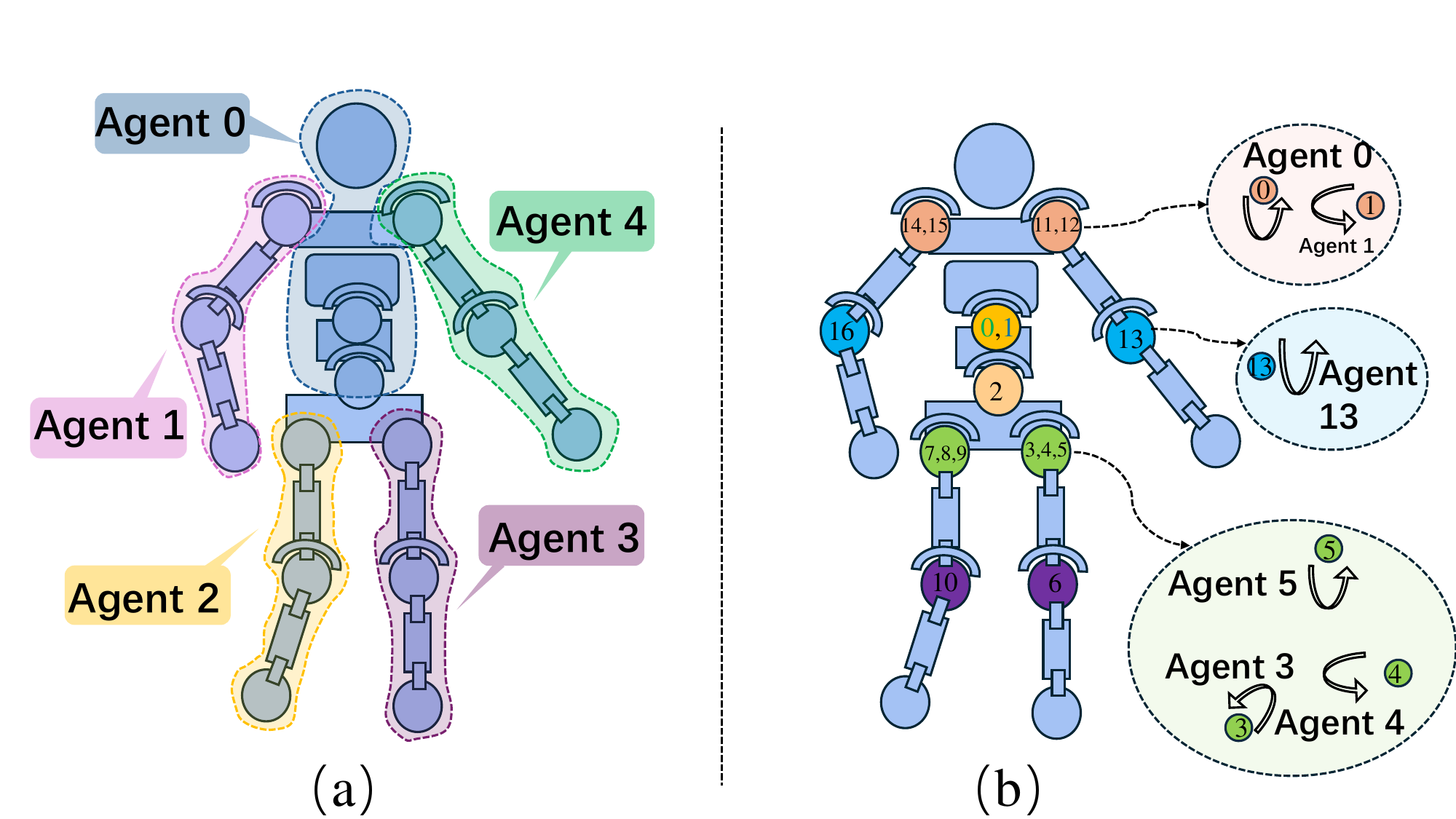}
    \caption{Limb- and JDoF-level multi-agent formulations for the humanoid robot. (a) Traditional limb-level formulation, where each limb is formulated as an independent homogeneous agent. (b) JDoF-level formulation, where each torque-actuated JDoF is formulated as an independent heterogeneous agent with physical attributes and functional roles.}  
    \label{fig:jdoF}
\end{figure}

To address these issues, some researchers introduce multi-agent reinforcement (MARL), where joints or limbs are treated as multiple homogeneous agents~\cite{lee2025learning, zhong2024heterogeneous}, as shown in Fig. \ref{fig:jdoF}(a). However, in fact, various joints in the human body exhibit obvious differences in muscle composition, movement directions, dimensions, and coupling~\cite{sferrazza2024body}. For HRs, their hip joints (high torque and heavy load) and elbow joints (high sensitivity and light load) also exhibit fundamental physical heterogeneity in dynamic characteristics, DoF, and roles in whole-body balance~\cite{sferrazza2024body}. Existing homogeneous MARL-based methods overlook heterogeneous attributes in joint and limbs and fail to learn effective actions for different agents with the same objectives but different decisions, generating incongruous motions.

Regarding the structure of the control model, some researchers adopt the end-to-end framework that directly maps observations and proprioceptions to control actions~\cite{sferrazza2024body, tevet2025closd, zhang2025hilo}.  These methods are simple and effective for humanoid robots with low DoF, but struggle to train an effective model in state-action spaces for HRs with high DoF. Other researchers employ a hierarchical model~\cite{figueroa2025fast} that separates high-level task planning from low-level joint control, where the high-level model often uses a large model combined imitation learning and the low-level model typically employs terminal controllers to perform planning. Nevertheless, these methods often decouple the learning of the high-level planning model and the low-level control models, often leading to insufficient coordination between high- and low-level models and stiff motions. In addition, the large model used in the high-level model often leads to a long inference time and poor real-time performance.

To address these issues, we propose a hierarchical heterogeneous multi-agent reinforcement learning ({H\textsuperscript{2}MARL}) for coordinating humanoid whole-body control, %, where the joint DoFs in HRs are formulated as heterogeneous agents following the characteristics of human joints, and a hierarchical coordinated control model that couples the high- and low-level controllers is developed to enable coordinated optimization between pose planning and joint tracking. In conclusion, 
which offers the following three contributions: 1) We propose a JDoF-level physical-heterogeneity-aware policy factorization. The humanoid whole-body control is formulated as a cooperative heterogeneous multi-agent problem at the joint-degree-of-freedom (JDoF) level, where each torque-actuated JDoF is conditioned on an explicit physical descriptor containing its actuation, kinematic, and dynamic properties, as shown in Fig. \ref{fig:jdoF}(b), providing a fine-grained inductive bias that enables different JDoFs to learn specialized control behaviors while maintaining coordination through centralized training. 2) We develop a hierarchical policy–executor coupling mechanism in which the JDoF-level pose planner and joint executor interact with the environment through the same closed-loop trajectory and are iteratively optimized rather than trained independently, preventing the distribution mismatch caused by freezing the executor. 3) Extensive experiments conducted in the MuJoCo and Isaac Gym simulator demonstrate that {H\textsuperscript{2}MARL} outperforms state-of-the-art methods in metrics for measuring coordination and human-like qualities. Through controlled comparisons of agent granularity, heterogeneous conditioning, and closed-loop collaborative optimization, we also isolate the contribution of each design and demonstrate that performance improvement comes from the proposed JDoF-level policy factorization and policy–executor coupling rather than from directly combining existing HAPPO, HST, and conventional PD components.

\section{Related Work}
\subsection{Whole-body Control for Humanoid Robots}
Current mainstream methods of humanoid robot WBC are based on large model and imitation learning~\cite{hu2023learning}, which can allow robots to achieve human-like locomotion and application in the fields of commercial services~\cite{li2025humanoid,gu2026humanoid}, home care~\cite{li2025humanoid}, entertainment performances~\cite{zhang2025hilo,tevet2025closd}, etc., but they heavily rely on a large number of expert demonstrations, limiting the applications in some hazardous industrial scenarios~\cite{li2025humanoid}. RL-based methods are increasingly being adopted by researchers due to the advantage of not requiring labeled data in recent years, and have demonstrated robust whole-body locomotion and loco-manipulation on humanoid and legged robots~\cite{zhou2025alarm, liu2025opt2skill, kumagai2025contact}. 
In particular, hierarchical RL is applied in robot control, where the controller is often divided into two layers, including a high-level planning model that generates abstract locomotion commands or selected skills and a low-level control model that executes specific actions, to enhance learning efficiency and the ability to solve long-range tasks~\cite{peng2017deeploco}. However, these methods often decouple the high-level planning model and the low-level control model, and the low-level model typically employs homogeneous coarse-grained skill executors, making it difficult to express the fine-grained heterogeneity and coupled coordination among joints in high-DoF HRs, resulting in insufficient coordination between high- and low-level models. Instead, we formulate the high-level controller as a set of heterogeneous agents and design a hierarchical policy–executor coupling mechanism to address the issues of homogeneity between joints and inconsistency between the two-level controllers, respectively.

\subsection{Heterogeneous Multi-Agent Systems}
Heterogeneous multi-agent systems (HMAS) have been applied in the fields of smart logistics, smart grids, and post-disaster search and rescue due to the ability to handle complex collaborative tasks~\cite{yan2024cooperative}. Researchers proposed some classic HMASs such as ROMA~\cite{wang2020roma}, RODE~\cite{wang2020rode}, and MPN~\cite{zhang2022markovian} to improve the collaboration performance of agents. However, these methods still follow the paradigm of sharing homogeneous parameters, which is similar to that in MAPPO~\cite{yu2022surprising} and IPPO, and only work well in tasks with isomorphic agents. Heterogeneous agent reinforcement learning (HARL) methods such as Heterogeneous-Agent PPO (HAPPO)~\cite{kuba2021trust} and HATD3~\cite{zhong2024heterogeneous} are proposed for effective collaboration among heterogeneous multi-agents. Some researchers use HARL to control various types of single-robot agents, such as quadrupedal robots and humanoid robots~\cite{yan2024cooperative}, performing high-dimensional actions, alleviating the difficulty of single-agent RL methods in precisely coordinating the motions. However, humanoid robots are not simple combinations of stacked limbs, and different parts of the body have different properties. These methods of unified formulation joints or limbs struggle to represent such fine-grained heterogeneous coupling characteristics at the joint level, leading to insufficient coordination. 

The most similar method to ours is HAPPO which solves heterogeneous-agent policy optimization, but does not specify how a humanoid's physically heterogeneous JDoFs should be factorized into agents nor how such policies should be coupled with a temporal executor. Instead, each torque-actuated DoF of joints is formulated as an independent heterogeneous agent according to the physically heterogeneous characteristics in our method, and the JDoF-level pose planner and joint executor interact with the environment through the same closed-loop trajectory, enhancing the coordination and human-like qualities.

\section{JDoF-Level Physical-Heterogeneity-Aware Policy Factorization}
Humanoid whole-body control is challenged by high-DoF coupling and joint-level physical heterogeneity, where joints differ in torque limits, inertia, degrees of freedom, and kinematic locations. The WBC methods based on homogeneous parameter sharing impose strong inductive bias on these heterogeneous control units, making it difficult to jointly achieve stability and agility and resulting in incongruous motions. Existing methods typically model the robot as a monolithic controller or adopt coarse limb-level agents. However, heterogeneity also exists within the degrees of freedom of a single joint. Although sharing the same physical body parts, these degrees of freedom differ in rotation axis, torque generation, roles in balance and locomotion, and coupling with adjacent joints. For example, sagittal hip motion mainly contributes to forward propulsion, while lateral hip motion focuses on balance and posture compensation. Formulating them as a single agent compresses multiple control roles into one policy, which is contrary to the structure composition of human bodies and increases optimization difficulty.

Motivated by the intrinsic heterogeneity of humanoid actuation dimensions, we propose a JDoF-
level physical-heterogeneity-aware policy factorization, where each torque-actuated JDoF is formulated as an independent heterogeneous agent with physical attributes and functional roles, as shown in Fig.~\ref{fig:jdoF}.

Let $N$ denote the number of JDoF agents. The agent set could be expressed by $\mathcal{I}=\{1,\ldots,N\}$. We formalize WBC as a cooperative Partially Observable Markov Game $\mathcal{G}=\langle \mathcal{S},\{\mathcal{O}_i\}_{i=1}^{N},\{\mathcal{A}_i\}_{i=1}^{N},P,r,\gamma\rangle$, where $s_t\in\mathcal{S}$ is the global state at the time step $t$. $o_{i,t}\in\mathcal{O}_i$ is the local observation, and $a_{i,t}\in\mathcal{A}_i$ is the action of Agent $i$ at the time step $t$. The joint action of all agents at the time step $t$ is $a_t=(a_{1,t},\ldots,a_{N,t})$. $P(s_{t+1}|s_t,a_t)$ is the transition function. $r$ and $\gamma$ are the shared reward and discount factor, respectively. According to HAPPO~\cite{kuba2021trust}, the finite-horizon objective $J(\pi)$ is defined as Eq. \ref{eq::HAPPOobj}.
\begin{equation}
J(\pi)=\mathrm{E}_{\pi,P}\left[\sum_{t=0}^{T-1}\gamma^t r(s_t,a_t)\right].
\label{eq::HAPPOobj}
\end{equation}

To formulate the heterogeneity, Agent $i$ is characterized by the descriptor $\xi_i$, as shown by Eq. \ref{eq::xi}.
\begin{equation}
\xi_i=
[c_i,u_i,\tau_i^{max},q_i^{min},q_i^{max},d_i,m_i],
\label{eq::xi}
\end{equation}
where $c_i$, $u_i$, $\tau_i^{max}$, $(q_i^{min},q_i^{max})$, $d_i$, and $m_i$ denote the joint type, actuation axis, torque limit, motion range, kinematic location, and inertia-related properties, respectively. These descriptor values are directly read from the used simulators and do not need to be set manually. These physical priors enable each agent to represent its own dynamics and functional role.

By leveraging the JDoF physical descriptor, we explicitly inject heterogeneous physical priors into the decentralized policy and transform the entire policy optimization problem through JDoF-level factorization, where JDoF-level factorization produces differentiated policies while preserving coordinated behavior. Therefore, the WBC policy $\pi_{\Theta}$ is formulated as a cooperative HMAS, where each JDoF-level agent learns an individual policy $\pi_{\theta_i}$, as shown in Eq. \ref{eq::HMARL}.

\begin{equation}
\pi_{\Theta}(a_t|o_t)=
\prod_{i=1}^{N}\pi_{\theta_i}(a_{i,t}|o_{i,t},\xi_i).
\label{eq::HMARL}
\end{equation}

Following HAPPO, each agent maintains an independent actor without parameter sharing, while a centralized critic enables cooperative optimization under the Centralized Training with Decentralized Execution framework.

\section{Hierarchical policy–executor coupling mechanism}

\begin{figure*}[tb!]
    \centering
    \includegraphics[width=0.9\textwidth]{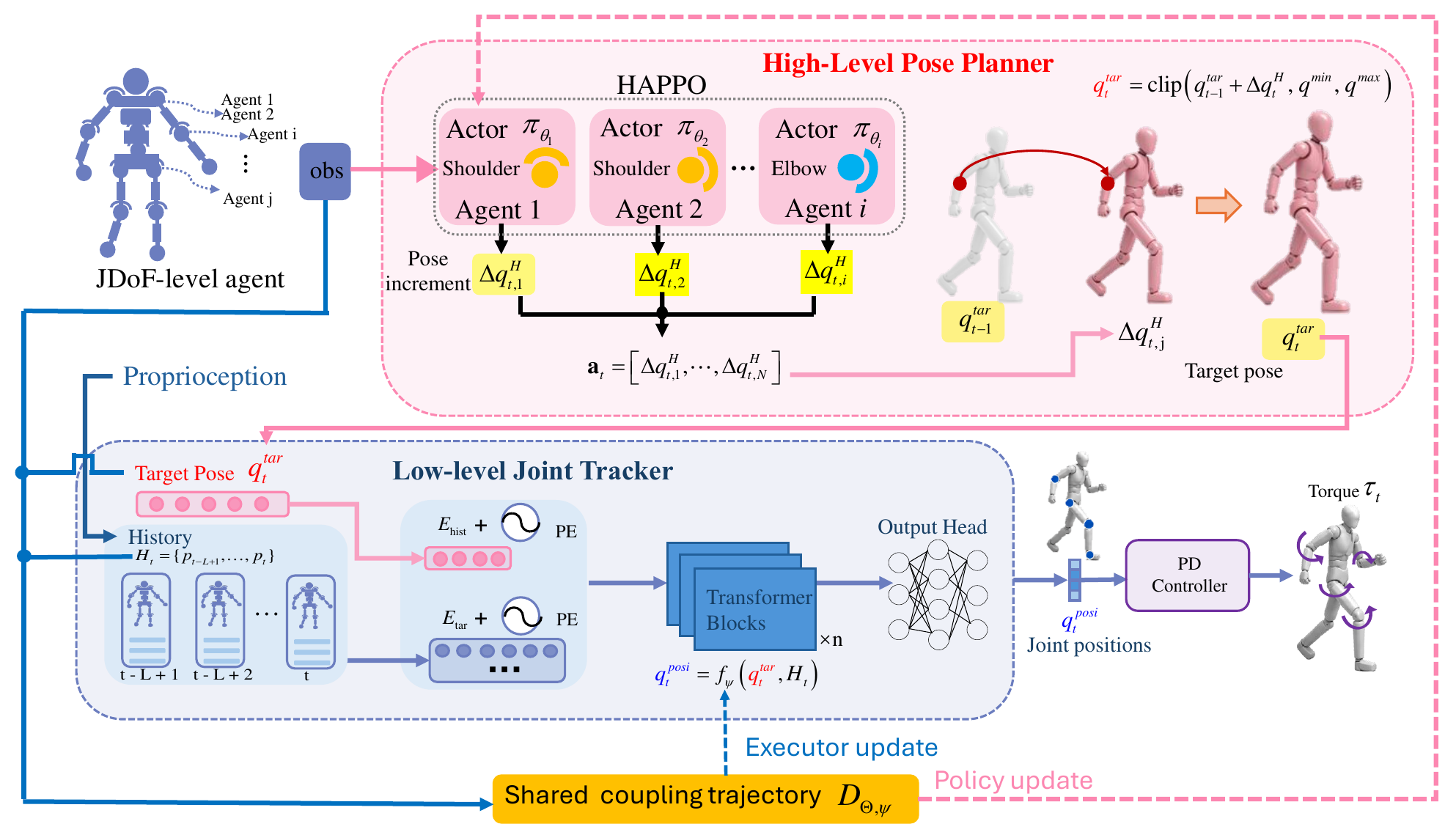}
    \caption{{H\textsuperscript{2}MARL} overview. The high-level planner generates JDoF-level pose increments and converts them into absolute target poses. The low-level joint executor receives the absolute target poses combined with historical proprioception and outputs executable joint positions which are converted to torques by a PD controller. Both the high-level policy and the low-level executor are optimized by the coupling trajectories in a closed-loop.}
    \label{fig:m2}
\end{figure*}

%To address the issue of insufficient coordination between the high- and low- level models in existing MARL-based methods for humanoid whole-body control, we propose a 
The proposed hierarchical policy–executor coupling mechanism consists of a high-level pose planner, a low-level joint executor, and a Closed-loop Policy–Executor Co-Optimization (CPECO) mechanism, as shown in Fig.~\ref{fig:m2}. At the high-level pose planner, each heterogeneous JDoF-level agent takes observation $o_{i,t}^{H}$ as input and outputs a pose increment $\Delta q_t^{H}$. The target pose is obtained by adding $\Delta q_t^{H}$ to the target pose at the previous time step $q_{t-1}^{tar}$. The low-level joint executor receives $q_t^{tar}$ and historical proprioception $H_t$ and outputs joint positions $q_t^{posi}$, which are converted to joint torques $\tau_t$ by a Proportional-Derivative (PD) controller. Both the high-level policy and the low-level executor are optimized by the coupling trajectories in a closed-loop.

\subsection{High-Level Pose Planner}
The high-level controller in existing hierarchical WBC methods typically generates absolute target poses of joints, which may be infeasible to track due to dynamic physical constraints and limited control bandwidth of the robot since the performance of the low-level joint executor is not considered in these methods, resulting in unstable joint position tracking and discontinuous target actions. To address this issue, we introduce a pose increment representation method to calculate joint target poses. Instead of predicting absolute poses, the high-level controller is formulated as a heterogeneous pose increment planner and generates adaptive joint target corrections with respect to the current target pose, improving joint action continuity while preserving task-driven adaptability. 

For each heterogeneous JDoF-level agent, the high-level observation is defined as:
\begin{equation}
o^{H}_{i,t}
=
[z^{local}_{i,t},z^{body}_{t},\xi_i,e_i],
\end{equation}
where $z^{local}_{i,t}$ contains the agent proprioception (position, velocity, and previous action). $z^{body}_{t}$ represents global body states. $\xi_i$ is the agent descriptor, and $e_i$ denotes the agent identity. The actor of each agent adopts a heterogeneous Gaussian policy optimized by HAPPO, as shown by Eq. \ref{eq::HAPPOpolicy}.
\begin{equation}
\pi_{\theta_i}^{H}
(\Delta q^{H}_{i,t}|o^{H}_{i,t})
=
\mathcal{N}
(\mu^{H}_{i,t},(\eta^{H}_{i,t})^2).
\label{eq::HAPPOpolicy}
\end{equation}
where $\mu_{i,t}^{H}$ and $\eta_{i,t}^{H}$ denote the Gaussian mean and standard deviation, respectively, and the increment correction is sampled by Eq. \ref{eq:q_H}.

\begin{equation}
\Delta q^{H}_{i,t}
=
\alpha_i
\tanh
(
\mu^{H}_{i,t}
+
\eta^{H}_{i,t}\odot\epsilon_{i,t}
),
\quad
\epsilon_{i,t}\sim\mathcal{N}(0,I).
\label{eq:q_H}
\end{equation}

The scaling factor $\alpha_i$ adapts the increment range according to the physical properties of the agents, as shown in Eq. \ref{eq::scalfactor}.

\begin{equation}
\alpha_i = \alpha \cdot \tau_i^{\max} \cdot (q_i^{\max} - q_i^{\min}) \cdot d_i / m_i
\label{eq::scalfactor}
\end{equation}
where $\alpha$ is a global scaling parameter that controls the overall scaling of the pose increments. $\alpha$ is fixedly set to 0.001 in our method.

This mechanism enables physically consistent increment generation, where JDoF-level agents can produce larger corrections, while stabilization-related JDoF-level agents focus on precise adjustments. During training, the actors in HAPPO are optimized by the cooperative objective, as shown in Eq. \ref{eq::HAPPOobj}. With a centralized critic, the actors of heterogeneous agent are allowed to learn specialized behaviors while maintaining whole-body coordination.

After obtaining increment corrections, the target pose of the high-level pose planner is calculated by Eq. \ref{eq:qtar}.

\begin{equation}
q^{tar}_{i,t}
=
clip
(
q^{tar}_{i,t-1}
+
\Delta q^{H}_{i,t},
q^{min}_i,
q^{max}_i
),
\label{eq:qtar}
\end{equation}
where the clipping operation guarantees joint physical limits. 

\subsection{Low-Level Joint Executor}
The high-level controller generates task-oriented target poses, but directly executing these commands is difficult due to the strong temporal dependency and dynamic constraints of humanoid motion. Therefore, a low-level controller is required to transform high-level motion intentions into physically feasible joint commands.

Unlike conventional position tracking methods that only consider the current target pose, humanoid motion execution requires historical states to capture motion evolution and contact dynamics to maintain the continuity of the motion. In this paper, we design a low-level joint executor based on the Human State Transformer (HST)~\cite{fu2024humanplus} as the temporal joint executor due to its ability to represent long-range temporal dependencies and generate context-aware joint trajectories.

To adapt the execution of target poses according to the current motion context and improve tracking smoothness, the proposed low-level joint executor takes the target pose $q_t^{tar}$ generated by the high-level planner, as well as the historical proprioception $H_t=\{p_{t-L+1},...,p_t\}$ as input, where $p_t$ contains torso orientations, base velocities, joint positions, velocities, and previous actions. $L$ is the step length of the historical proprioception. The executable joint position $q_t^{posi}$ is generated by Eq. \ref{eq:q_set}.

\begin{equation}
q_t^{posi}=f_{\psi}(q_t^{tar},H_t),
\label{eq:q_set}
\end{equation}
where $f_{\psi}$ denotes the temporal execution network in HST, which is optimized by a joint objective $\mathcal{L}_{hst}$ including the tracking accuracy term that is derived from the difference of the current joint position and the current target pose, the temporal smoothness term that is derived from the difference of the current and previous joint positions, and the torque regularization term $\tau_t$ that is from the PD controller.%, as shown by Eq. \ref{eq:hst_loss}.

\begin{equation}
\mathcal{L}_{hst}
=
\mathrm{E}
[
w_q\|q^{posi}_{t}-q^{tar}_{t}\|_2^2
+
w_s\|q^{posi}_{t}-q^{posi}_{t-1}\|_2^2
+
w_{\tau}\|\tau_t\|_2^2
],
\label{eq:hst_loss}
\end{equation}
where $w_q$, $w_s$, and $w_{\tau}$ are the weights of the three terms. In this paper, they are set to 1.0, 0.1, and 0.001, respectively, following the settings in HST.

Finally, $q_t^{posi}$ is fed into the PD controller to generate torque commands for WBC of the robot, where the proportional and derivative gains are fixedly set to $K_p{=}100$ and $K_d{=}5$, respectively, in our method.

\subsection{Closed-loop Policy–Executor Co-Optimization}
The high- and low-level models in existing hierarchical models for whole-body control of HRs often use the method of independent optimization, that is, the low-level model is first trained, and the parameters of the low-level model are frozen when the high-level model is then trained. However, during the RL training process in our method, the target action distribution generated by the high-level policy continuously changes as the policy is updated. If the low-level controller is frozen, its policy distribution will gradually deviate from the action distribution generated by the high-level policy, leading to insufficient inter-layer coordination and policy performance decline. Therefore, we design a closed-loop policy–executor co-optimization for the proposed {H\textsuperscript{2}MARL}, where the high-level policy and the low-level executor are jointly optimized by the coupling trajectories in a closed-loop.

At each time step $t$, the policy of the high-level planner generates JDoF-level pose increments according to Eq. ~\ref{eq:q_H}, which are converted into target poses by Eq. ~\ref{eq:qtar}. The low-level HST then outputs joint positions based on the targets and historical proprioceptions, as shown in Eq. ~\ref{eq:q_set}. The resulting interaction trajectories $D_{\Theta,\psi}$ are shared by both levels.
\begin{equation}
D_{\Theta,\psi}
=
\{
s_t,o_t^H,q_t^{tar},H_t,q_t^{posi},\tau_t,r_t,s_{t+1}
\}_{t=0}^{T-1},
\end{equation}
where $\Theta$ and $\psi$ denote the parameters in the high- and low-level models, respectively. The trajectory is not merely shared for data reuse. It establishes an optimization coupling between the planner and executor. %The high-level heterogeneous actors are updated with HAPPO and a centralized critic. Meanwhile, the low-level policy minimizes execution errors under the target poses generated by the high-level planner. 
The optimization process is summarized in Algorithm ~\ref{alg:HMARL-WBC}.

%It is important to clarify that H²MARL does not claim HAPPO, HST, PD control, or pose-increment parameterization as individual algorithmic innovations. These components serve as established optimization, temporal-execution, torque-interface, and action-parameterization backbones, respectively. The methodological novelty of H²MARL lies instead in (i) reformulating humanoid whole-body control as a JDoF-level physically heterogeneous policy factorization with explicit embodiment descriptors, and (ii) introducing a closed-loop policy–executor co-optimization mechanism that couples high-level policy learning with low-level executable dynamics. The controlled ablation studies are designed to isolate these two mechanisms from the adopted components.

\begin{algorithm}[t]
\caption{Closed-loop Policy–Executor Co-Optimization}
\label{alg:HMARL-WBC}
\begin{algorithmic}[1]

\REQUIRE Environment $\mathcal{E}$, high-level policy $\pi_{\theta}^{H}$ parametered by $\theta$,
low-level joint executor $f_{\psi}$ parametered by $\psi$, centralized critic $Q_{\phi}$ parametered by $\phi$, trajectories $D_{\Theta,\psi}$
\ENSURE Optimized $\pi_{\theta}^{H}$ and $f_{\psi}$

\STATE Initialize parameters $\theta$, $\psi$, and $\phi$

\FOR{each training iteration $k$}

\STATE Collect trajectories $D_{\Theta,\psi}^k$ by the hierarchical model

\FOR{each time step $t$}

\STATE Generate JDoF-level pose increments by Eq.~\ref{eq:q_H}

\STATE Construct target poses by Eq.~\ref{eq:qtar}

\STATE Generate executable joint positions by Eq.~\ref{eq:q_set}

\STATE Obtain $\tau_t$ and interact with $\mathcal{E}$ by a PD controller

\ENDFOR

\STATE Estimate advantages using $Q_{\phi}$

\STATE Update $\pi_{\theta}^{H}$ by Eq.~\ref{eq::HAPPOobj}

\STATE Fix $\pi_{\theta}^{H}$ and update  $f_{\psi}$ by minimizing Eq.~\ref{eq:hst_loss}

\STATE Fix  $\pi_{\theta}^{H}$ and $f_{\psi}$, and update $Q_{\phi}$ using $D_{\Theta,\psi}^k$

\ENDFOR

\end{algorithmic}
\end{algorithm}

\section{Experiments}
\subsection{Experimental Setup}
We conducted experiments on the Humanoid-v2 walking task in the MuJoCo~\cite{todorov2012mujoco} and Isaac Gym~\cite{makoviychuk2021isaacgymhighperformance} benchmarks. %on which most SOTA methods such as MAPPO~\cite{yu2022surprising} and HATD3~\cite{zhong2024heterogeneous} evaluate their performance. Following baselines ~\cite{yu2022surprising, zhong2024heterogeneous}, at the beginning of each episode, the humanoid robot is initialized with a random orientation and required to move to a predefined goal direction. An episode ends when the robot falls or the defined termination conditions are met.
Following the standard configurations in MuJoCo and Gymnasium, the robot contains 10 joints with a total of 17 actuated joint torques, and each of them is modeled as a heterogeneous agent.% The action is defined as a vector $\tau_t \in \mathbf{R}^{17}$, and each element in $\tau_t$ corresponds to an actuated joint torque, subject to environment-specified limits. 
Each agent provides a proprioception $s_t \in \mathbf{R}^{376}$ at time step $t$, including joint-related features, rigid-body dynamics, and external contact forces, etc.

To improve locomotion quality, we introduce two task-specific auxiliary reward terms in our method, including a slip penalty for suppressing undesired foot sliding during ground contact and an alternation reward for encouraging coordinated left--right stepping. The overall reward at time step $t$ is formulated by Eq. \ref{eq:overall_reward}.
\begin{equation}
\label{eq:overall_reward}
\begin{array}{rcl}
R_t &=& r_t^{\mathrm{task}}
+\lambda_{\mathrm{tr}}r_t^{\mathrm{tra}}
+\lambda_{\mathrm{st}}r_t^{\mathrm{sta}}
+\lambda_{\mathrm{alt}}r_t^{\mathrm{alt}}   \\
&&-\lambda_{\mathrm{slip}}r_t^{\mathrm{slip}}   
-\lambda_{\tau}\|\tau_t\|_2^2
-\lambda_{\mathrm{sm}}
\|q_t^{\mathrm{posi}}-q_{t-1}^{\mathrm{posi}}\|_2^2, 
\end{array}
\end{equation}
where $r_t^{\mathrm{task}}$ is the task reward in HAPPO. The tracking and stability rewards $r_t^{\mathrm{tra}}$ and $r_t^{\mathrm{sta}}$, which encourage accurate pose tracking and stable torso posture, respectively, also follow the settings in HAPPO. The designed alternation reward $r_t^{\mathrm{alt}}$ promotes coordinated left--right stepping, while the designed slip reward $r_t^{\mathrm{slip}}$ suppresses tangential foot motion during ground contact. The last two terms penalize excessive joint torques and abrupt changes in joint positions, respectively. The coefficients $\lambda_{\mathrm{tr}}$, $\lambda_{\mathrm{st}}$, $\lambda_{\mathrm{alt}}$, and $\lambda_{\mathrm{slip}}$ balance the rewards of pose tracking, torso stability, gait alternation, and slip suppression, respectively, and are set to $\lambda_{\mathrm{slip}}{=}0.01$ and $\lambda_{\mathrm{alt}}{=}0.05$ for the two auxiliary terms. $\lambda_{\tau}$ and $\lambda_{\mathrm{sm}}$ are the torque and motion-smoothness regularization, respectively. All coefficients are set uniformly in our experiments and do not require fine-tuning.

We compare {H\textsuperscript{2}MARL} with some SOTA methods, including HAPPO~\cite{kuba2021trust}, MAPPO~\cite{yu2022surprising}, and HATD3~\cite{zhong2024heterogeneous}. %MAPPO is a representative on-policy cooperative MARL method based on centralized training, and HATD3 is an off-policy method designed for heterogeneous-agent continuous control. 
For fairness, all methods use the same configurations and experimental settings as HAPPO.
We use four metrics to measure the coordination and human-like qualities of humanoid control, including: 1) Foot Slip (FS) ~\cite{radosavovic2024real} that quantifies the undesired relative motion between a supporting foot and the ground, 2) Torso Stability (TS) that measures deviations in upper-body posture during locomotion, 3) Gait Symmetry (GS) ~\cite{li2025humanoid, radosavovic2024real} that evaluates the consistency between the left and right legs and bilateral coordination, and 4) Motion Smoothness (MS) ~\cite{radosavovic2024real} that is used to assess motion continuity and naturalness. 

Following the baseline HAPPO, the policy is trained for 10 million environment steps. All experiments are conducted on an NVIDIA GeForce RTX 4090 GPU with 24 GB memory and an Intel Xeon Gold 6530 CPU with 30 cores.

\subsection{Comparative Experiments with SOTA Methods}
As shown in Table~\ref{tab:sota_comparison}, {H\textsuperscript{2}MARL} achieves the best results in metrics including FS, TS, and GS, demonstrating superior foot--ground contact stability, upper-body balance, and bilateral coordination. In particular, it outperforms HAPPO and HATD3 in all four metrics, indicating a clear overall advantage and more coordinating whole-body motions over existing homogeneous and heterogeneous MARL baselines. Compared to MAPPO, the most competitive method, {H\textsuperscript{2}MARL} still improves FS, TS, and GS by 7.3\%, 35.0\%, and 14.8\%, respectively, indicating more reliable ground contact, better posture regulation, and more coordinated left--right leg movements. This is attributed to JDoF-level heterogeneous formulation that enables individual agent to learn specialized policies according to their physical properties and functional roles, as well as closed-loop policy–executor co-optimization that reduces the mismatch between high- and low-level models. It is noted that {H\textsuperscript{2}MARL} exhibits a little higher MS than MAPPO, indicating that the motion smoothness is slightly inferior to MAPPO. This is because JDoF-level heterogeneous formulation introduces more frequent joint-level adjustments to enhance stability and coordination, which reduces motion smoothness to a certain extent.

\begin{table}[tb!]
\centering
\small
\setlength{\tabcolsep}{3.5pt}
\renewcommand{\arraystretch}{1.1}
\caption{Comparative Results with SOTA methods.}
\label{tab:sota_comparison}
\begin{tabular}{lccccc}
\hline
Method  & FS $\downarrow$ & TS $\downarrow$ & GS $\uparrow$ & MS $\downarrow$ \\
\hline
HAPPO  & 0.9827 & 0.5091 & 0.000150 & 7.3819 \\
MAPPO  & 0.5790 & 0.3873 & 0.000905 & \textbf{0.2329} \\
HATD3  & 0.8220 & 0.4739 & 0.000050 & 2.3818 \\
{H\textsuperscript{2}MARL}(ours)  & \textbf{0.5367} & \textbf{0.2516} & \textbf{0.001039} & 0.3514 \\
\hline
\end{tabular}

\end{table}

\subsection{Ablation Studies}

\subsubsection{Ablation study for different mechanisms}
We compare the complete {H\textsuperscript{2}MARL} with the following four variants: i) {H\textsuperscript{2}MARL} with the heterogeneous agent formulation being replaced by the homogeneous agent formulation in MAPPO (\textit{w/o HA}), ii) {H\textsuperscript{2}MARL} with the closed-loop policy–executor co-optimization being replaced by independent optimization of the high- and low-level models (\textit{w/o CPECO}), iii) {H\textsuperscript{2}MARL} where the designed slip reward and alternation reward in our method are removed from Eq. \ref{eq:overall_reward} (\textit{w/o Rew}), and iv) {H\textsuperscript{2}MARL} with neither the HA nor CPECO as well as removing the designed two reward terms (\textit{w/o All}). In order to verify coordination and human-like performance in different environments, we conducted ablation studies on another benchmark, Isaac Gym ~\cite{makoviychuk2021isaacgymhighperformance}. Following the standard in this benchmark, Velocity Error (VE), Cost of Transport (CoT), Motion Jerk (MJ), Centroidal Angular Momentum (CAM), and Torso Stability (TS) are adopted as metrics. The results are reported in Table~\ref{tab:ablation}. 
The complete {H\textsuperscript{2}MARL} achieves the best performance in four metrics. \textit{w/o CPECO} shows substantial increases in CoT, MJ, and CAM compared with the complete model, demonstrating that  closed-loop policy–executor co-optimization is essential for energy-efficient and stable locomotion. \textit{w/o CPECO} achieves the lowest TS, since the reduced torso motion comes at the expense of overall locomotion quality, as evidenced by the significant degradation in other four metrics. \textit{w/o HA} causes a low but consistent performance degradation across all metrics, indicating that explicitly formulating joint heterogeneity improves control accuracy and execution quality. \textit{w/o Rew} consistently degrades overall performance, demonstrating the designed reward is essential to guide the training process. %\textit{{H\textsuperscript{2}MARL} w/o All} performs worst in VE, CoT, and MJ and also shows clear degradation in CAM and TS. 

\begin{table}[t]
\centering
\small
\setlength{\tabcolsep}{3.5pt}

\caption{Ablation study of different mechanisms on Isaac Gym.}
\label{tab:ablation}
\begin{tabular}{lccccc}
\hline
Method & VE$\downarrow$ & CoT$\downarrow$ & MJ$\downarrow$ & CAM$\downarrow$ & TS$\downarrow$ \\
\hline
H\textsuperscript{2}MARL      & \textbf{0.4807} & \textbf{1.5213} & \textbf{0.7448} & \textbf{1.0324} & 0.0140 \\
w/o HA                        & 1.0360 & 5.8804 & 2.9680 & 3.8187 & 0.0417 \\
w/o CPECO                       & 0.9908 & 11.9586 & 6.4986 & 6.0132 & \textbf{0.0106} \\
w/o Rew             & 1.0580 & 14.2529 & 2.6578 & 12.9126 & 0.1691 \\
w/o All (HAPPO)             & 1.3667 & 14.7716 & 19.5962 & 9.0912 & 0.0333 \\
\hline
\end{tabular}

\end{table}

\subsubsection{Evaluation for JDoF-level agent formulation}
To investigate the effect of the granularity of agent formulation, we compare the JDoF-level formulation with following formulation methods: i) single-agent formulation where a single agent controls the whole body, ii) body part (BP)-level formulation where the lower half body including the hip and legs is treated as an agent and the upper half body is treated as another agent; iii) limb-level formulation where each agent is modeled using each limb in the robot. As shown in Table~\ref{tab:partition}, JDoF-level formulation reduces VE, CoT, MJ, CAM, and TS by 71.4\%, 79.3\%, 88.3\%, 26.7\%, and 71.7\%, respectively, compared with the limb-level formulation method, demonstrating that JDoF-level formulation improves task performance and control efficiency while enabling more stable posture regulation and more effective distribution of control responsibilities. BP-level formulation achieves intermediate performance. Although it achieves the lowest CAM, the coarser agent partition limits fine-grained joint coordination, resulting in inferior overall locomotion performance compared with JDoF-level formulation. Single-agent formulation also performs substantially worse than JDoF-level formulation, further demonstrating the importance of explicitly modeling JDoF-level heterogeneity.

\begin{table}[t]
\centering
\setlength{\tabcolsep}{3.5pt}
\small

\caption{Comparison of different agent formulation strategies.}
\label{tab:partition}
\begin{tabular}{lccccc}
\hline
Method & VE$\downarrow$ & CoT$\downarrow$ & MJ$\downarrow$ & CAM$\downarrow$ & TS$\downarrow$ \\
\hline
Single-agent & 1.4003 & 10.0510 & 2.8654 & 0.6105 & 0.0307 \\
BP-level & 0.9695 & 3.5571 & 2.1089 & \textbf{0.5330} & 0.0539 \\
Limb-level & 1.6811 & 7.3547 & 6.3729 & 1.4076 & 0.0495 \\
JDoF-level & \textbf{0.4807} & \textbf{1.5213} & \textbf{0.7448} & 1.0324 & \textbf{0.0140} \\
\hline
\end{tabular}

\end{table}

% \subsubsection{Evaluation for pose increment}
% To verify the advantages of using the pose increment as the agent action rather than the absolute pose, We compare these two formulations under identical training and evaluation settings. The experimental results are shown in Fig.~\ref{fig:bar_residual_fullpose}. Compared to the absolute-pose formulation, the pose-increment formulation reduces VE, CoT, MJ, CAM and TS by 52.7\%, 61.5\%, 73.0\%, 65.9\%, and 27.8\%, respectively, indicating fewer tracking errors, higher control efficiency, and smoother motion execution, better torso stability.%, demonstrating that the pose increment formulation provides a more effective representation for generating coordinated and physically feasible humanoid motions.

% \begin{figure}[tb!]
%     \centering
%     \includegraphics[width=0.4\textwidth]{photo/Residual_vs_FullPose.png}
%     \caption{Comparison of pose increment and absolute pose.}
%     \label{fig:bar_residual_fullpose}
% \end{figure}

\subsubsection{Evaluation of different rewards}
The slip reward and the alternation reward in Eq. \ref{eq:overall_reward} are also the differences between our method and the baselines. To analyze the effect of the two reward terms, we compare the complete reward function with the three variants by separately removing each term  and removing both. The results are shown in Table~\ref{tab:reward_ablation}. The complete reward function (Complete) achieves the best performance. Compared to the complete reward, removing the slip reward (\textit{w/o Slip}) increases VE, CoT, MJ, CAM, and TS by 100.4\%, 595.9\%, 972.5\%, 550.8\%, and 87.9\%, respectively, indicating a larger velocity error, higher energy consumption, greater body oscillations, and worse motion stability.%, which highlights the importance of suppressing foot slip during locomotion. 
Removing the alternation reward (\textit{w/o Alt}) also increases VE, CoT, MJ, CAM, and TS by 66.0\%, 503.4\%, 974.1\%, 422.0\%, and 25.7\%, respectively, indicating that encouraging coordinated left--right stepping substantially improves locomotion accuracy, control efficiency, motion smoothness, and body stability. The method removing both reward terms performs the worst of all. 
%When both reward terms are removed (\textit{w/o Both}), the overall performance further degrades, further demonstrating the significance of these two rewards.

\begin{table}[t]
\centering
\setlength{\tabcolsep}{3.5pt}
\small

\caption{Ablation study of different reward terms.}
\label{tab:reward_ablation}
\begin{tabular}{lccccc}
\hline
Method & VE$\downarrow$ & CoT$\downarrow$ & MJ$\downarrow$ & CAM$\downarrow$ & TS$\downarrow$ \\
\hline
Complete & \textbf{0.4807} & \textbf{1.5213} & \textbf{0.7448} & \textbf{1.0324} & \textbf{0.0140} \\
w/o Slip & 0.9632 & 10.5866 & 7.9879 & 6.7192 & 0.0263 \\
w/o Alt & 0.7981 & 9.1799 & 7.9996 & 5.3886 & 0.0176 \\
w/o Both & 1.0580 & 14.2529 & 2.6578 & 12.9126 & 0.1691 \\
\hline
\end{tabular}

\end{table}

 \subsection{Real-time Performance Analysis}
%Real-time performance is one of the most important factors in decision-making systems. In this paper, 
We calculate the 50th percentile value (P50), the 99th percentile value (P99), and the average value of the single-step running time for 2000 steps as the metrics to evaluate the real-time performance. As shown in Table. \ref{tab:realtime}, {H\textsuperscript{2}MARL} introduces only a slight computational overhead compared to HAPPO. The high-level planner requires an average inference time of 8.332 ms, which is only 0.655 ms higher than HAPPO. This increase is mainly due to the additional JDoF-level heterogeneous formulation and closed-loop policy–executor co-optimization. 
The total inference time of {H\textsuperscript{2}MARL} is 9.259 ms, with a frequency of more than 100 Hz, which satisfies the real-time requirement of most HR control systems.

\begin{table}[t]
\centering
\small
\setlength{\tabcolsep}{3.5pt}

\caption{Single-step inference time (ms) for comparative methods.}
\label{tab:realtime}
\begin{tabular}{lccccc}
\hline
Method & P50 & P99 & Mean & Std \\
\hline
HAPPO & 7.589 & 8.896 & 7.677 & 0.315 \\
MAPPO & 4.958 & 6.209 & 5.005 & 0.426 \\
HATD3 & 7.533 & 8.353 & 7.592 & 0.381 \\   
{H\textsuperscript{2}MARL} (High-level) & 8.241 & 9.609 & 8.332 & 0.330 \\
{H\textsuperscript{2}MARL} (Low-level) & 0.910 & 1.100 & 0.927 & 0.225 \\
\hline
\end{tabular}

\end{table}

\section{Conclusions}
We proposed {H\textsuperscript{2}MARL}, a hierarchical heterogeneous multi-agent reinforcement learning method for humanoid whole-body control in this paper. Each torque-actuated JDoF in the robot is formulated as an independent heterogeneous agent to allow the policy to understand the differences in physical properties and functional roles among joints. A hierarchical policy–executor coupling mechanism is also developed to improve coordination between high- and low-level models, enhancing coordination of robot motion. Experimental results show that {H\textsuperscript{2}MARL} achieves better locomotion stability, gait coordination, and motion naturalness than SOTA methods.

\bibliographystyle{IEEEtran}
\bibliography{aaai2026}

\end{document}